\documentclass[conference]{IEEEtran}

\usepackage{cite}

\ifCLASSINFOpdf
  \usepackage[pdftex]{graphicx}
  \graphicspath{{./figures/}}
\else
\fi
\ifCLASSOPTIONcompsoc
  \usepackage[caption=false,font=normalsize,labelfont=sf,textfont=sf]{subfig}
\else
  \usepackage[caption=false,font=footnotesize]{subfig}
\fi
\begin{document}
\title{Towards Dependability Metrics for Neural Networks}

\author{\IEEEauthorblockN{Chih-Hong Cheng\IEEEauthorrefmark{1}, Georg N\"{u}hrenberg\IEEEauthorrefmark{1}, Chung-Hao Huang\IEEEauthorrefmark{1}, Harald Ruess\IEEEauthorrefmark{1} and Hirotoshi Yasuoka\IEEEauthorrefmark{2}}
	\IEEEauthorblockA{\IEEEauthorrefmark{1}fortiss - Research Institute of the Free State of Bavaria \\
		Email: \texttt{\{cheng,nuehrenberg,huang,ruess\}@fortiss.org}}
	\IEEEauthorblockA{\IEEEauthorrefmark{2}DENSO CORPORATION\\
		Email: \texttt{hirotoshi\_yasuoka@denso.co.jp}}
	}

\maketitle

\begin{abstract}
Artificial neural networks (NN) are instrumental in realizing highly-automated driving functionality. An overarching challenge  is to identify best safety engineering practices for NN and other learning-enabled components. In particular, there is an urgent need for an adequate set of metrics for measuring all-important NN dependability attributes.
We address this challenge by proposing a number of NN-specific and efficiently computable metrics for measuring NN dependability attributes including robustness, interpretability, completeness, and correctness.
\end{abstract}

\IEEEpeerreviewmaketitle

\section{Introduction}
\vspace{-0.5mm}

Artificial neural networks (NN) are instrumental in realizing a number of important features  in safety-relevant applications such as highly-automated driving. 
In particular, vision-based perception, the prediction of drivers' intention, and even end-to-end autonomous control are usually based on NN technology.
State-of-the-practice safety engineering processes (cmp. ISO 26262) require that safety-relevant components, including NN-enabled ones, demonstrably satisfy their respective safety goals. 

Notice that the transfer of traditional testing methods  and corresponding test coverage metrics such as MC/DC (cmp. DO 178C) to NN  may lead to an exponential (in the number of neurons) number of branches to be investigated ~\cite{cheng2018neural}\@. Such an exponential blow-up is not practical as typical NN may be comprised of millions of neurons.
Moreover, a straightforward adaptation of structural coverage metrics for NN, e.g., the percentage of activated neurons for a given test set~\cite{pei2017deepxplore}, does not take into account that the activation of single neurons is usually not strongly connected to the result of the whole network. The challenge therefore is to develop a set of NN-specific and efficiently computable metrics for measuring various aspects of the dependability of NN.

In previous work we have been generating test cases for NN testing based on finite partitions of the input space and by relying on  predefined sets of application-specific scenario attributes~\cite{cheng2018quantitative}\@.
Besides correctness and completeness of NN we also identified robustness~\cite{cheng2017maximum} and interpretability~\cite{cheng2018neural} as 
important NN dependability attributes.

Here we build on our previous work on testing NN, and we propose a set of metrics for measuring the  $\textbf{RICC}$  dependability attributes of NN, which are informally described as follows.
\begin{itemize}
    \item \textbf{R}obustness of a NN against various effects such as distortion or adversarial perturbation (which is closely related to  \emph{security}). 
    \item \textbf{I}nterpretability in terms of understanding important aspects of what a NN has actually learned. 
    \item \textbf{C}ompleteness in terms of ensuring that the data used in training has possibly covered all important scenarios. 
    \item \textbf{C}orrectness in terms of a NN able to perform the perception task without errors. 
\end{itemize}

\begin{figure}[t]
\includegraphics[width=0.8\columnwidth,page=2,clip,trim=16mm 0 16mm 0]{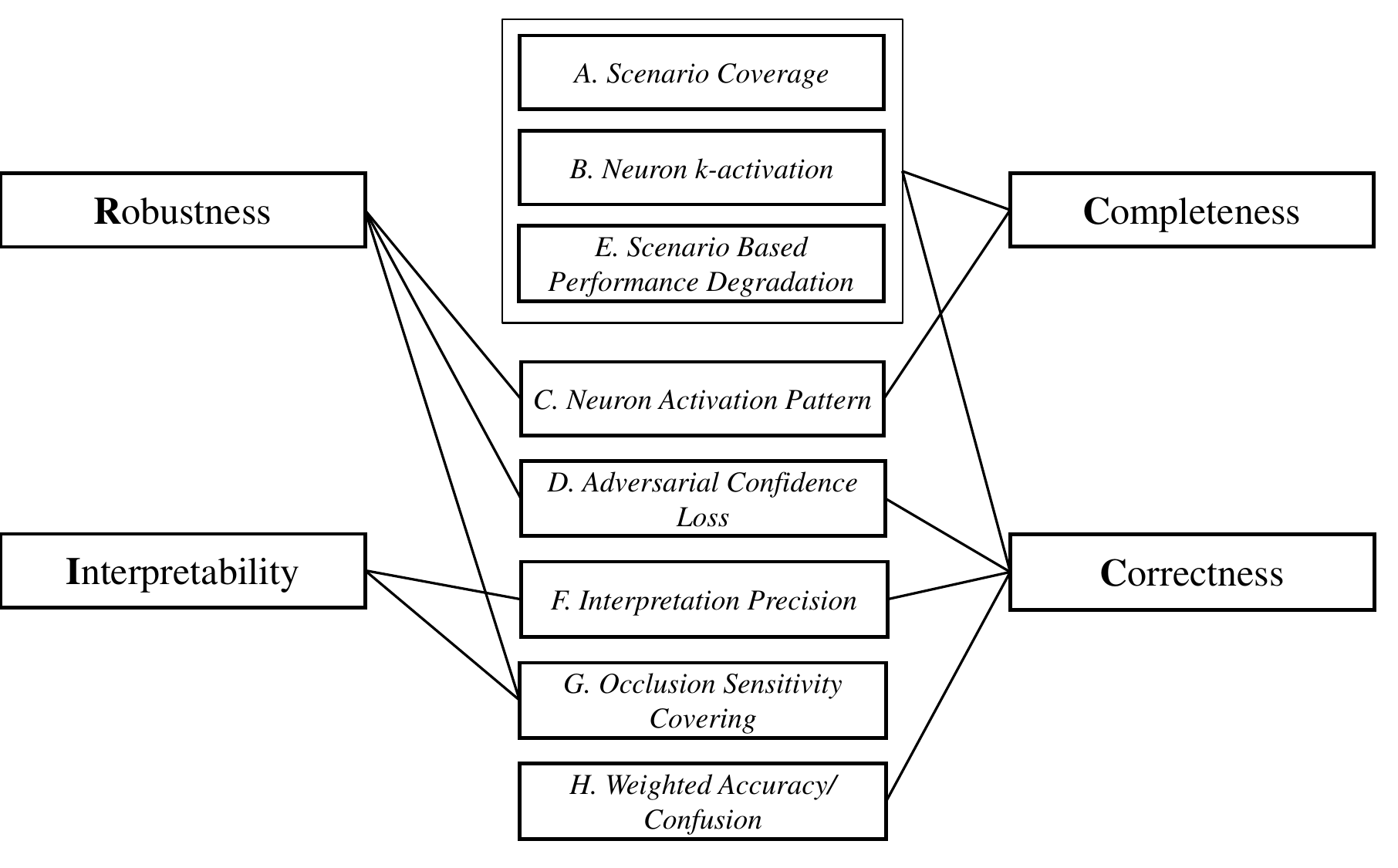}
\centering

\vspace{-3mm}
\caption{Relations between \textbf{RICC} criteria and the proposed metrics. The group of metrics~\textit{A., B.} and \textit{E.} cover completeness and correctness.}
\label{fig.metric.inputs}
\end{figure}

The main contribution of this paper is a complete and efficiently computable set of NN-specific metrics for measuring  $\textbf{RICC}$  dependability attributes. 
Fig.~\ref{fig.metric.inputs} illustrates how the metrics cover the space of \textbf{RICC}, where at least two metrics relate to each criterion.

\vspace{-2mm}
\section{Quality Metrics}
\vspace{-1mm}

\subsection{Scenario coverage metric  $\mathcal{M}_{scene}$}
\vspace{-1mm}
Similar to the class imbalance problem~\cite{japkowicz2002class} when training classifiers in machine learning, one needs to account for the presence of all relevant scenarios in training datasets for NN for autonomous driving. 
A scenario over a list of $\mathcal{C}= \langle C_1, \ldots, C_n \rangle$ of operating conditions (e.g., weather and road condition) is given by a valuation of each condition. E.g., let $C_1  = \{sunny, cloudy, rainy\}$ represent the weather condition, $C_2 = \{stone, mud, tarmac\}$ represent the road surfacing, and $C_3  = \{straight, curvy\}$ represent the incoming road orientation. Then $(sunny, stone, straight)$ and $(rainy, tarmac, curvy)$ constitute two possible scenarios.

Since for realistic specifications of operating conditions, checking the coverage of all scenarios is infeasible due to combinatorial explosion, 
our proposed \emph{scenario coverage metric} is based on the concept of $2$-\emph{projection} and is tightly connected to the existing work of combinatorial testing, covering arrays and their quantitative extensions~\cite{lawrence2011survey,nie2011survey,cheng2018quantitative}. 

\begin{figure}[t]
\includegraphics[width=0.9\columnwidth]{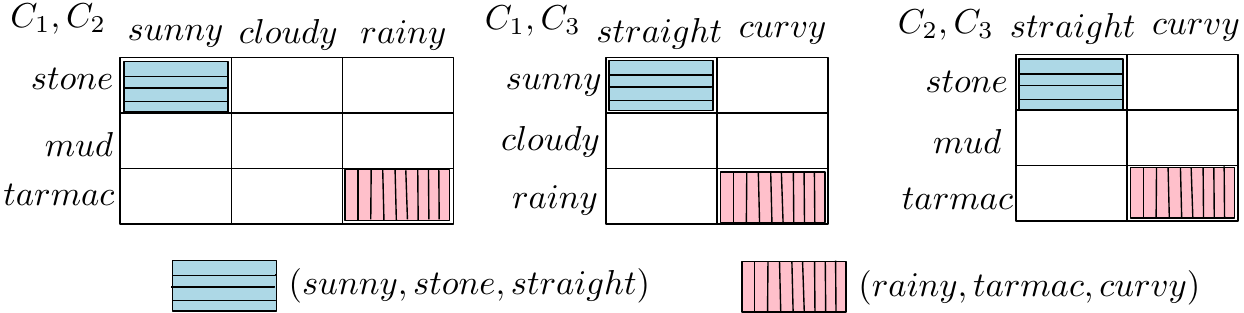}
\centering
\vspace{-3mm}
\caption{Computing scenario coverage metric via $2$-projection table}
\label{fig.2Projection.coverage}
\end{figure}

\subsubsection*{Assumption} Computing the scenario coverage metric requires that the dataset is semantically labeled according to the specified operating conditions, such that for each data point it can be determined whether it belongs to a certain scenario.

\vspace{1mm}
\subsubsection*{Computing  $\mathcal{M}_{scene}$ } 

The  metric starts by preparing a table recording all possible \emph{pairs} of operating conditions, followed by iterating each data point to update the table with occupancy. Lastly, compute the ratio between occupied cells and the total number of cells. Eq.~\ref{eq.M.scene} summarizes the formula, and an illustration can be found in Fig.~\ref{fig.2Projection.coverage}, where a dataset of two data points $\{(sunny, stone, straight), (rainy, tarmac, curvy)\}$  achieves  $\mathcal{M}_{scene}=\frac{2+2+2}{9+6+6}$. 
    
    \vspace{-3mm}
 \begin{equation}
 \label{eq.M.scene}
        \mathcal{M}_{scene} := \frac{\mathrm{\#\ of\ cells\ occupied \ by\ the\ data\ set}}{\mathrm{\#\ of\ cells\ from\ }  2\mathrm{-projection\ table}}
\end{equation}

Provided that for each $C_i$, the size of $C_i$ is bounded by constant $\alpha$ (i.e., the categorization is finite and discrete), then the denominator can at most be ${n \choose 2}\alpha^2$, i.e., the number of data points required for full coverage is \emph{polynomially} bounded.

\subsubsection*{Relations to \emph{\textbf{RICC}} \& improving $\mathcal{M}_{scene}$}

The metric reflects completeness and correctness attributes of \textbf{RICC}. To improve the metric,  one needs to discover new scenarios. For the example in Fig.~\ref{fig.2Projection.coverage}, an image satisfying the scenario $(cloudy,mud,curvy)$ can efficiently increase the metric from $\frac{2+2+2}{9+6+6}$ to $\frac{3+3+3}{9+6+6}$.

\subsection{Neuron $k$-activation metric $\mathcal{M}_{neu-k-act}$}
\vspace{-1mm}
The previously described input space partitioning may also be observed by the activation of neurons.
By considering  ReLU activation as an indicator of successfully detecting a feature, 
for close-to-output layers where high-level features are captured, the combination of neuron activation in the same layer also forms scenarios (which are independent from the specified operating conditions).
Again, we encounter combinatorial explosion, e.g., for a layer of $256$ neurons, there is a total of $2^{256}$ scenarios to be covered.
Therefore, similar to the 2-projection in the scenario coverage metric, this metric  only monitors whether the input set has enabled all activation patterns for every neuron pair or triple in the same layer. 

\subsubsection*{Assumption} The user specifies an integer constant~$k$  and a specific layer to be analyzed. Assume that the layer has $c$ neurons. 

\subsubsection*{Computing  $\mathcal{M}_{neu-k-act}$ } 

The  metric starts by preparing a table recording all possible $k$-\emph{tuples} of on-off activation for neurons in the layer being analyzed (similar to Fig.~\ref{fig.2Projection.coverage} with each $C_i$ now having only $1$ and $0$ status), followed by iterating each data point to update the table with occupancy. The denominator is given by the number of cells, which has value ${c \choose k}(2^{k})$.
    
    \vspace{-3mm}
 \begin{equation}
        \mathcal{M}_{neu-k-act} := \frac{\mathrm{\#\ of\ occupied\ cells\ due\ to\ the\ data\ set}}{{c \choose k}(2^{k})}
\end{equation}    

Note that when $k=1$, our defined neuron $k$-activation  metric $\mathcal{M}_{neu-k-act}$ subsumes commonly seen neuron coverage acting over a single layer~\cite{neuroncoverage,pei2017deepxplore}, where one analyzes the on-off cases for each individual neuron. 

\subsubsection*{Relations to \emph{\textbf{RICC}} \& improving $\mathcal{M}_{neuron-k-act}$}

The metric reflects the completeness and correctness attribute of \textbf{RICC}. To improve the metric,  one needs to provide inputs that allows enabling different neuron activation patterns.

\subsection{Neuron activation pattern metric $\mathcal{M}_{neu-pattern}$}
\vspace{-1mm}
Encountering the combinatorial explosion, while $k$-activation metric captures the completeness, our designed \emph{neuron activation pattern metric} is used to understand the distribution of activation. For inputs within the same scenario, intuitively the activation pattern should be similar, implying that the number of activated neurons should be similar. 

\subsubsection*{Assumption} 

The user provides an input set $\textsf{In}$, where
all images belong to
the same scenario,
and specifies a layer of the NN (with $c$ neurons) to be analyzed.
Furthermore, the user chooses the number of groups~$\gamma$, for a partition of~$\textsf{In}$ into~$\gamma$ groups $G_1(\textsf{In}), \ldots, G_{\gamma}(\textsf{In})$, where for group~$G_i(\textsf{In})$, $i\in\{1, \ldots, \gamma\}$, the number of activated neurons in the specified layer is within the range $[\frac{c}{\gamma}(i-1), \frac{c}{\gamma}(i)]$ for each input in this group. 

\subsubsection*{Computing  $\mathcal{M}_{neu-pattern}$}

Let $G_{j}(\textsf{In})$ be the largest set among $G_1(\textsf{In}), \ldots, G_{\gamma}(\textsf{In})$. Then the metric is evaluated by considering all inputs whose activation pattern, aggregated using the number of neurons being activated,  significantly deviates from the majority.  

\vspace{-3mm}
 \begin{equation}
        \mathcal{M}_{neu-pattern} := \frac{\sum_{i\;:\;i\not\in \{j-1,j,j+1\} } |G_{i}(\textsf{In})| }{|\textsf{In}|}
\end{equation}

\subsubsection*{Relations to \emph{\textbf{RICC}} \& improving $\mathcal{M}_{neu-pattern}$}

This metric reflects the robustness and completeness attribute of \textbf{RICC}, as well as interpretability. To improve the metric, one requires careful examination over the reason of diversity in the activation pattern under the same scenario. 

\subsection{Adversarial confidence loss metric $\mathcal{M}_{adv}$}
\vspace{-1mm}
Vulnerability w.r.t.\ adversarial inputs~\cite{szegedy2013intriguing} is an important quality attribute of NNs, which are used for image processing and designed to be used in safety-critical systems.
As providing a formally  provable guarantee against all possible adversarial inputs is hard, our proposed adversarial confidence loss metric is useful in providing engineers an estimate of how robust a NN is.

\subsubsection*{Assumption}

Computing $\mathcal{M}_{adv}$ requires that there exists a list of input transformers $\langle T_1, \ldots, T_n \rangle$ where for each~$T_i, (i\in \{1, \ldots, n\}) $,
given a parameter $\epsilon$ specifying the allowed perturbation, one derives a new input $\textsf{in'} = T_i(\textsf{in}, \epsilon)$ by transforming input~$\textsf{in}$.
Each $T_i$ is one of the known image perturbation techniques ranging from simple rotation, distortion, to advanced techniques such as FGSM~\cite{goodfellow2014explaining} or deepfool~\cite{moosavi2016deepfool}.

\subsubsection*{Computing  $\mathcal{M}_{adv}$}

Given a test set $\textsf{In}$, a predefined perturbation bound $\epsilon$, and the list of input transformers, let $\textsf{NN}(\textsf{in})$, where $\textsf{in} \in \textsf{In}$, be the output of the NN being analyzed, with larger value being better\footnote{Here the formulation also assumes that there exists a single output for the NN, but the formulation can be easily extended to incorporate multi-output scenarios.}. The following equation computes the adversarial perturbation loss metric.  

\vspace{-4mm}
\begin{equation}
        \mathcal{M}_{adv} := \frac{\sum_{\textsf{in} \in \textsf{In}} \textsf{min}_{i \in \{1, \ldots, N\}} \textsf{NN}(T_i(\textsf{in}, \epsilon)) - \textsf{NN}(\textsf{in})}{|\textsf{In}|} 
\end{equation}

\begin{figure}[t]
\subfloat[A vehicle image and three perturbed images. The largest classification performance drop is achieved by the FGSM technique.\label{fig.metric.adv}]{\includegraphics[width=0.49\columnwidth]{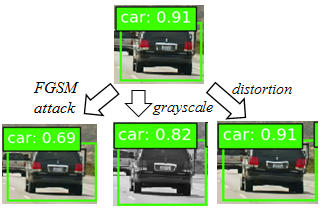}}
\hfill
\subfloat[This heatmap for a pedestrian contains nine hot pixels in orange, 30 occluding pixels and five hot and occluding pixels.\label{fig.interpretation.human}
]{\makebox[.49\columnwidth][c]{\includegraphics[width=0.2\columnwidth]{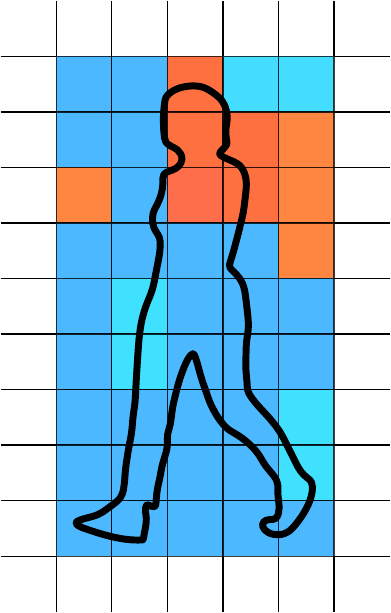}}}
\caption{Illustrating $\mathcal{M}_{adv}$ and a heatmap for $\mathcal{M}_{interpret}$ and $\mathcal{M}_{OccSen}$.}
\label{fig.metric.adv.and.heatmap}
\end{figure}

Intuitively,  $\mathcal{M}_{adv}$ analyzes the change of output value for input $\textsf{in}$ due to a perturbation $(\textsf{NN}(T_i(\textsf{in}, \epsilon)) - \textsf{NN}(\textsf{in}))$, and selects one which leads to largest performance drop among all perturbation techniques, i.e., it makes the computed value of $\textsf{NN}(T_i(\textsf{in}, \epsilon)) - \textsf{NN}(\textsf{in})$ most negative.
A real example is shown in Fig.~\ref{fig.metric.adv}, 
where the FGSM attack yields the largest classification performance drop among three perturbation techniques, which changes the probability of car from $0.91$ to $0.69$. Thus, the largest negative value of the probability difference $\textsf{NN}(T_i(\textsf{in}, \epsilon)) - \textsf{NN}(\textsf{in})$ for this image is~$-0.22$.
Lastly, average the computed value over all inputs being analyzed. 



\subsubsection*{Relations to \emph{\textbf{RICC}} \& improving $\mathcal{M}_{adv}$}

The metric has a clear impact on robustness and correctness. To improve the metric,  one needs to introduce perturbed images into the training set, or apply alternative training techniques with provable bounds~\cite{kolter2017provable}. 

\vspace{-1mm}
\subsection{Scenario based performance degradation metric}

Here we omit details, but for commonly seen performance metrics such as validation accuracy or even quantitative statistic measures such as MTBF, one may perform detailed analysis by either considering each scenario, or by discounting the value due to missing input scenarios (the discount factor can be taken from the computed scenario coverage metric).

\vspace{-1mm}
\subsection{Interpretation precision metric $\mathcal{M}_{interpret}$}
The interpretation precision metric is intended to judge if a NN for image classification  or object detection makes its decision on the correct part of the image.
E.g., the metric can reveal that a certain class of objects is mostly identified by its surroundings, maybe because it only exists in similar surroundings in the training and validation data. 
In this case, engineers should test whether this class of object can also be detected in different contexts.

\subsubsection*{Assumption} 
For computing this metric,
we need a validation set that has image segmentation ground truth in addition to the ground truth classes (and bounding boxes), e.g., as in VOC2012 data set~\cite{pascal-voc-2012}.
\subsubsection*{Computing  $\mathcal{M}_{interpret}$ }
\begin{figure}
\centering
    \subfloat[Result of object detection]{\includegraphics[width=.5\columnwidth]{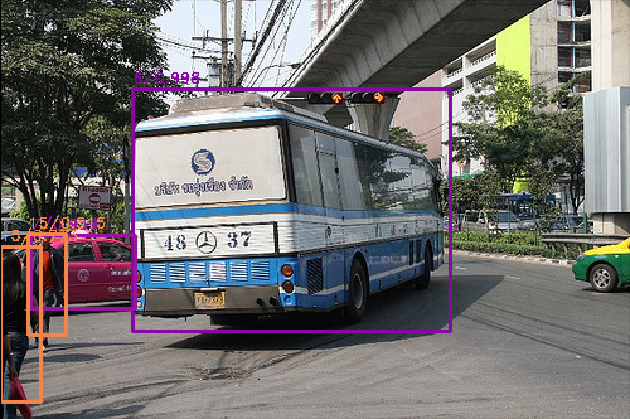}} \vspace{-4mm}\\
    \subfloat[Heatmap for red car (bottom left)] {\includegraphics[width=.495\columnwidth]{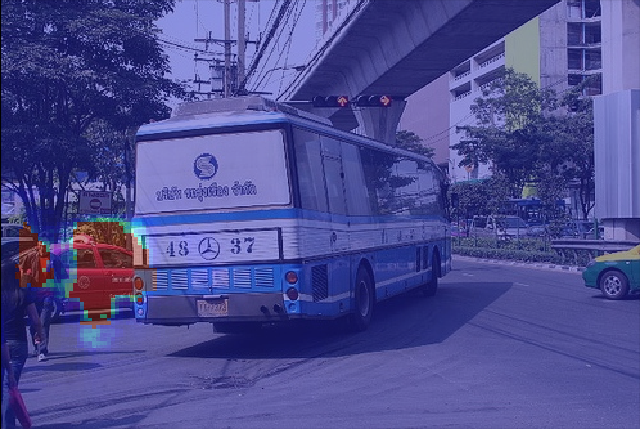}} \hfill
    \subfloat[$\mathcal{M}_{interpret}$ for $\rho$]{\includegraphics[width=.495\columnwidth,clip,trim=4mm 4mm 4mm 4mm]{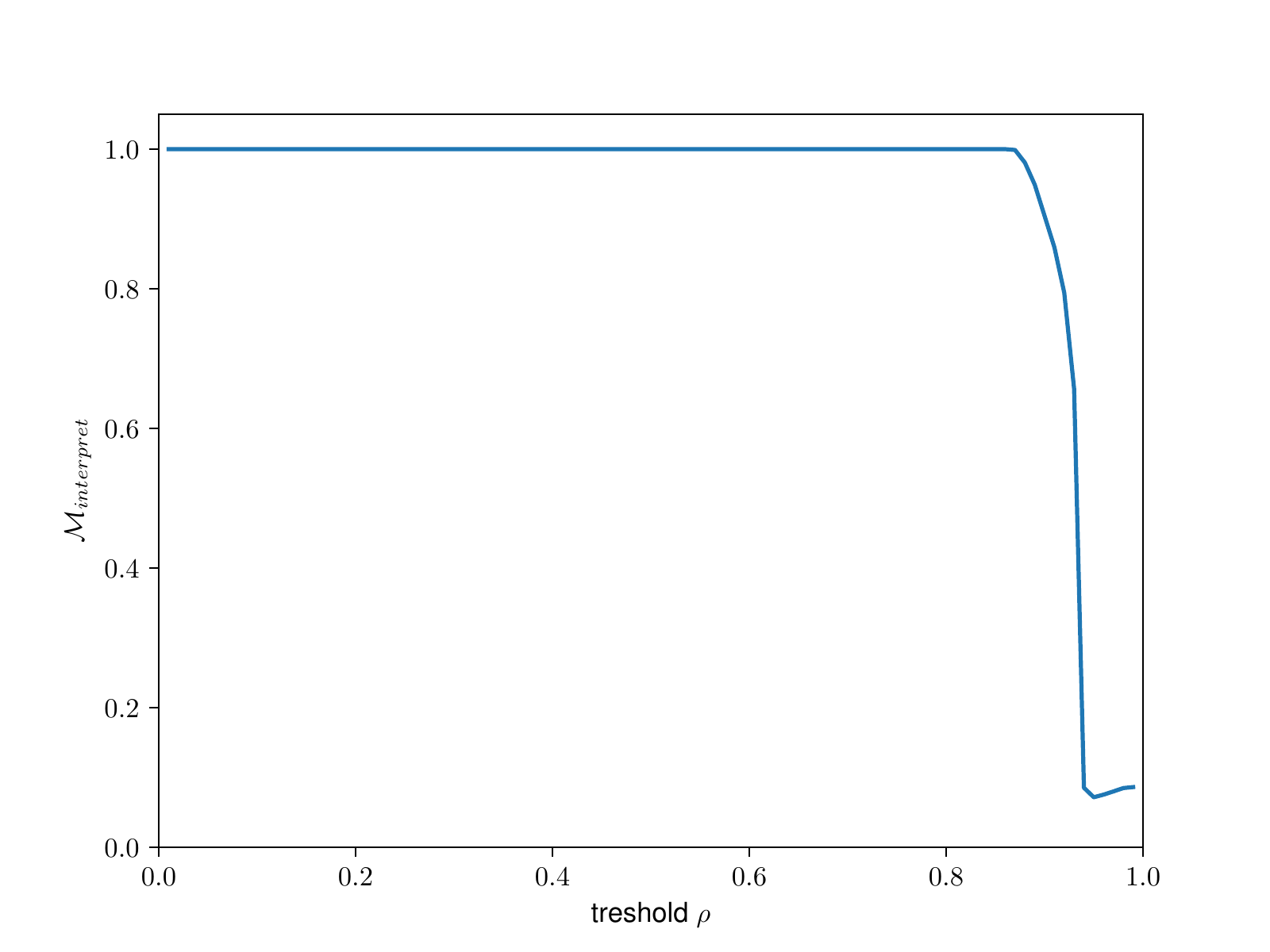}}\vspace{-4mm} \\
    \subfloat[Heatmap for the right person]{\includegraphics[width=.495\columnwidth]{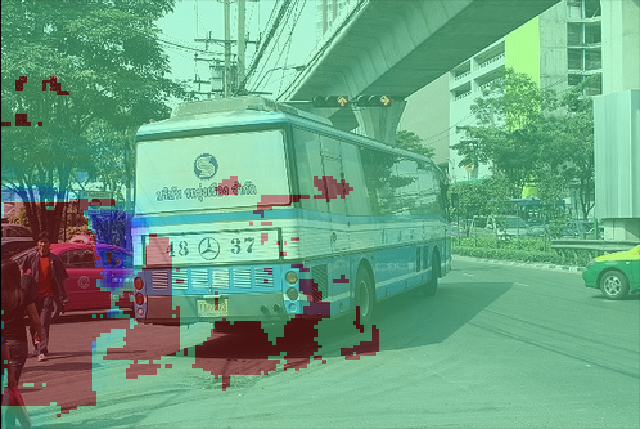}} \hfill
    \subfloat[$\mathcal{M}_{interpret}$ for $\rho$]{\includegraphics[width=.495\columnwidth,clip,trim=4mm 4mm 4mm 4mm]{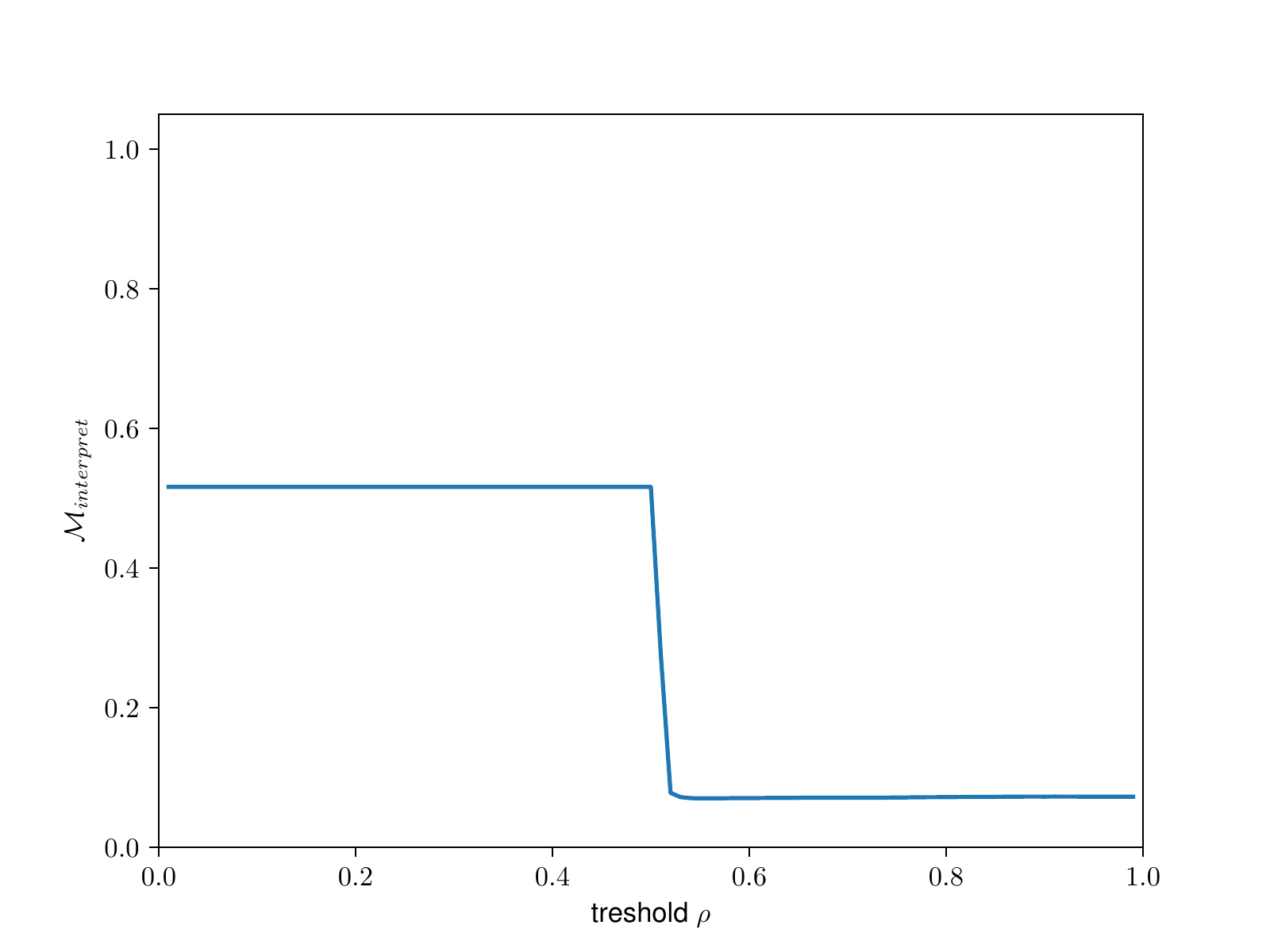}} 
    \caption{Computing $\mathcal{M}_{interpret}$ for red car and the right person in front of the red car. The metric shows that the red car is mostly identified by the correct areas. On the other hand, for the person there are a lot of hot pixels in incorrect regions.}
    \label{fig:interpretation_metric}
\end{figure}

Here we describe how the metric can be computed for a single detected object, where one can extend the computation to a set of images by posing average or min/max operators. A real example demonstrating the exact computation is shown in Fig.~\ref{fig:interpretation_metric}.
\begin{enumerate}
    \item Run the NN on the image to classify an object with probability $p$ (and obtain a bounding box
    in the case of object detection).
    \item Compute an occlusion sensitivity heatmap $H$, where each pixel of the heatmap $h\in H$ maps to a position of the occlusion on the image~\cite{zeiler2014visualizing}. The value of $h$ is given by the probability of the original class for the occluded image. For object detection we take the maximum probability of the correct class over all detected boxes that have a significant Jaccard similarity with the ground truth bounding box.
    \item For given probability threshold $\rho$ that defines the set of hot pixels as $P_{hot}=\{h\in H\ |\ h<\rho\}$ and the set of pixels that partly occlude the segmentation ground truth, denoted by $P_{occluding}$, the metric is computed as follows:
\end{enumerate}

\vspace{-7mm}
\begin{equation}
    \;\;\;\;\mathcal{M}_{interpret} = \frac{|P_{hot}\cap P_{occluding}|}{|P_{hot}|}
\end{equation}

An illustrative example of computing $\mathcal{M}_{interpret}$ can be found in Fig.~\ref{fig.interpretation.human}, where for the human figure only five out of nine hot pixels intersect the region of the human body. Thus $\mathcal{M}_{interpret}= \frac{5}{9}$. The set of thirty pixels constituting the human forms $P_{occluding}$.

\vspace{1mm}
\subsubsection*{Relations to \emph{\textbf{RICC}} \& improving $\mathcal{M}_{interpret}$}
The interpretation precision metric contributes to the interpretability and correctness of the \textbf{RICC} criteria. It may reveal that a NN uses a lot of context to detect some objects, e.g., regions surrounding the object or background of the image. In this case, adding images where these objects appear in different surroundings can improve the metric.

\vspace{-1mm}
\subsection{Occlusion sensitivity covering metric $\mathcal{M}_{OccSen}$}
\vspace{-1mm}
This metric measures the fraction of the object that is sensitive to occlusion.
Generally speaking, it is undesirable to have a significant probability drop if only a small part of the object is occluded.

Furthermore, care should be taken about the location of the occlusion sensitive area. If a certain part of an object class is occlusion sensitive in many cases (e.g., the head of a dog) it should be tested if the object can still be detected when this part is occluded (e.g., head of a dog is behind a sign post).
$\mathcal{M}_{OccSen}$ is computed in a similar way and based on the same inputs as $\mathcal{M}_{interpret}$:

\begin{enumerate}
    \item Perform steps 1) and 2) and determine $P_{hot}$ and $P_{occluding}$ as for $\mathcal{M}_{interpret}$.
    \vspace{1mm}
    \item Derive $\mathcal{M}_{OccSen} := \frac{|P_{hot}\cap P_{occluding}|}{|P_{occluding}|}$.
\end{enumerate}

If the value is high it indicates that many positions of small occlusions can lead to a detection error.
A low value indicates that there is a greater chance of still detecting the object when it is partly occluded.
An illustrative example of computing $\mathcal{M}_{OccSen}$ can be found in Fig.~\ref{fig.interpretation.human}, where for the human figure the heatmap only contains five hot pixels intersecting the human body (the head). As there are 30 pixels intersecting the region of the human, we have $\mathcal{M}_{OccSen}= \frac{5}{30}$.

\subsubsection*{Relations to \emph{\textbf{RICC}} \& improving $\mathcal{M}_{OccSen}$}
Occlusion sensitivity coverage covers the robustness and interpretability of \textbf{RICC}. If the metric values are too high for certain kinds of objects, an approach to improve it is to augment the training set with more images where these objects are only partly visible.

\vspace{-1mm}
\subsection{Weighted accuracy/confusion metric $\mathcal{M}_{confusion}$}
In object classification, not all errors have the same severity, e.g., confusing a pedestrian for a tree is more critical than in the opposite way. Apart from pure accuracy measures, one may employ fine-grained analysis such as specifying  penalty terms as weights to capture different classification misses. 

As such a technique is standard in performance evaluation of machine learning algorithms, the specialty will be how the weights of confusion are determined. Table~\ref{table.penality.terms} provides a summary over penalties to be applied in traffic scenarios, by reflecting the safety aspect. Misclassifying a pedestrian (or bicycle) to be background image (i.e., no object exists) should be set with highest penalty, as pedestrians are unprotected and it may easily lead to life threatening situations. 

\begin{table}[t]
\centering
\caption{Qualitative severity of safety to be reflected as weights}
\label{table.penality.terms}
\vspace{-2mm}
\begin{tabular}{|l|c|c|c|}\hline
 A is classified to B & B (pedestrian)  & B (vehicle) & B (background)   \\\hline
 A (pedestrian) & n.a. (correct) & $++$  &  $++++$  \\\hline
 A (vehicle) & $+$ & n.a. (correct) & $+++$    \\\hline
 A (background) & $+$ & $+$ & n.a. (correct)  \\\hline
\end{tabular}
\end{table}

\vspace{1mm}
\subsubsection*{Relations to \emph{\textbf{RICC}} \& improving $\mathcal{M}_{confusion}$}

The metric is a fine-grained indicator on  correctness. To improve the metric, either one trains the network with more examples, or one modifies the loss function such that it is aligned with the weighted confusion, e.g., it sets higher penalty term when misclassifying a ``pedestrian" to ``background".  

\vspace{-1mm}
\section{Outlook}
We propose a set of NN-specific and efficiently computable metrics for measuring the RICC dependability attributes of NN. 
At this point, we have also implemented a NN testing tool for evaluating the usefulness of our proposed set of metrics in on-going industrial NN developments. Our ultimate goal is to obtain a complete and validated set of NN dependability metrics. In this way, corresponding best practices can be identified as the basis of new safety processes for engineering NN-enabled components and systems.

\bibliographystyle{IEEEtran}


\end{document}